\documentclass{article}
\usepackage{spconf, amsmath, graphicx}
\usepackage{subfigure}
\usepackage{amssymb,bm}
\usepackage{mathrsfs}
\usepackage{booktabs}
\usepackage{multirow}
\usepackage{float}
\usepackage{textcomp}
\usepackage{color}
\usepackage{threeparttable}

\title{Region Proposal Network with Graph Prior and IoU-Balance Loss for Landmark Detection in 3D Ultrasound} 
\name{\parbox{\linewidth}{\centering Chaoyu Chen$^{1,2,\dagger}$, Xin Yang$^{1,2,\dagger}$, Ruobing Huang$^{1,2}$, Wenlong Shi$^{1,2}$, Shengfeng Liu$^{1,2}$, Mingrong Lin$^{1,2}$, \textit{Yuhao Huang$^{1,2}$, Yong Yang$^{1,2}$, Yuanji Zhang$^{3}$, Huanjia Luo$^{3}$, Yankai Huang$^{3}$, Yi Xiong$^{3}$, Dong Ni$^{1,2,*}$}} \thanks{$\dagger$ Authors contributed equally.} \thanks{* Corresponding author: \textit{nidong@szu.edu.cn}.} \thanks{The work in this paper was supported by the grant from National Key R${\&}$D Program of China (No. 2019YFC0118300), Shenzhen Peacock Plan (No. KQTD2016053112051497, KQJSCX20180328095606003), Medical Scientific Research Foundation of Guangdong Province, China (No. B2018031).}}
\address{$^{1}$National-Regional Key Technology Engineering Laboratory for Medical Ultrasound, \\Guangdong Key Laboratory for Biomedical Measurements and Ultrasound Imaging, \\School of Biomedical Engineering, Health Science Center, Shenzhen University, Shenzhen, China \\$^{2}$Medical UltraSound Image Computing (MUSIC) Lab, Shenzhen University, Shenzhen, China\\
$^{3}$Department of Ultrasound, Luohu People's Hosptial, Shenzhen, China}

\begin{document}
	%
	\maketitle
	\begin{abstract}
		3D ultrasound (US) can facilitate detailed prenatal examinations for fetal growth monitoring. To analyze a 3D US volume, it is fundamental to identify anatomical landmarks of the evaluated organs accurately. Typical deep learning methods usually regress the coordinates directly or involve heatmap-matching. However, these methods struggle to deal with volumes with large sizes and the highly-varying positions and orientations of fetuses. In this work, we exploit an object detection framework to detect landmarks in 3D fetal facial US volumes. By regressing multiple parameters of the landmark-centered bounding box (B-box) with a strict criteria, the proposed model is able to pinpoint the exact location of the targeted landmarks. Specifically, the model uses a 3D region proposal network (RPN) to generate 3D candidate regions, followed by several 3D classification branches to select the best candidate. It also adopts an IoU-balance loss to improve communications between branches that benefits the learning process. Furthermore, it leverages a distance-based graph prior to regularize the training and helps to reduce false positive predictions. The performance of the proposed framework is evaluated on a 3D US dataset to detect five key fetal facial landmarks. Results showed the proposed method outperforms some of the state-of-the-art methods in efficacy and efficiency. \par
	\end{abstract}	
	\begin{figure}[htb]
		\centering
		\includegraphics[width=1.0\linewidth]{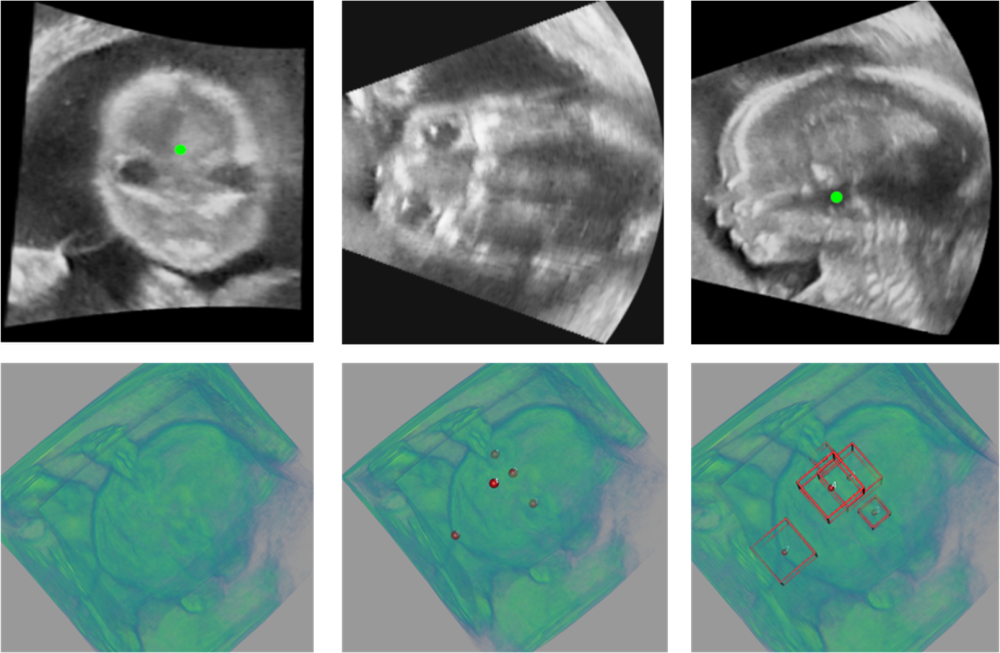}
		\caption{3D US of fetal face. First row: from left to right is the coronal, traverse and sagittal US plane of fetal face. Second row: from left to right is volume rendering of fetal face, 5 fetal facial landmarks and 5 landmark-centered B-boxes.}
		\label{fig:chalenges_box}
		\vspace{-0.3cm}
	\end{figure}\par
	\section{Introduction}
	As a low cost, real-time and non-radiation imaging modality, ultrasound (US) screening is ubiquitous in maternal and prenatal care. Compared with traditional 2D US scans, a 3D US  volume can provide a more complete record of anatomical information by offering a larger field of view. Detecting anatomical landmarks within the 3D US volumes can pave the way for plenty of automated applications in 3D US, such as the biometric measurement and volume registration \cite{huang2018vp,liu2019deep}. \par
	
	However, it is not a trivial task. The poor image quality, the presence of speckle and acoustic shadows,  and the large volume size of 3D US scans raise problems in designing  automated localization methods (Fig. \ref{fig:chalenges_box}). Large pose and size variability of fetuses further increases the difficulty of the task. An ideal landmark detection method needs to localize the target landmarks while recognize each of their classes simultaneously. Some of	the existing methods resort to directly regress the spatial coordinates of landmarks \cite{payer2016regressing,sofka2017fully,wang2019joint}. Others used heatmap-matching approaches. For example, Payer et al. \cite{payer2016regressing} proposed a deep neural network to predict the Gaussian heatmap of each landmark. In \cite{huang2018vp}, Huang et al. proposed to formulate the landmark detection as a segmentation task by directly segmenting landmark-centered bounding box (B-box) regions.  While being efficient, these models may struggle in learning both the localization and classification (recognize different landmarks) tasks using the same set of parameters, especially in 3D US. Furthermore, these regression methods only provide  prediction with the highest confidence, without providing alternatives for further correction. \par 
	In this work, we propose to explore object detection framework for landmark detection in 3D US. A landmark is determined by the position and the class identity of a B-box to be detected. The motivations behind this design are: \textit{1)} object detection method explicitly splits the localization and classification tasks with two branches; This design is more flexible and reduces the difficulties in each branch. Also, regressing multiple parameters of the landmark-centered B-boxes puts more constraint on learning than only predicting the landmark coordinates. \textit{2)} this framework allows post-processing and further evaluation. It can generate multiple candidates with different class scores that can be used in further refinement. \par
	There are abundant literature in object detection. Multi-stage detectors, like Faster R-CNN \cite{ren2015faster}, have found their applications in 2D US images \cite{lin2019multi}. These frameworks often achieved relatively high accuracy but might have compromises in efficiency. One-stage solution, like SSD \cite{liu2016ssd}, can significantly reduce time costs while it may sacrifice accuracy. Properly tackling this trade-off is crucial to 3D medical image analysis, where the input size is large while the target object is small. A very recent work by Xu et al. \cite{xu2019efficient} shared similar concern and implemented an efficient 3D Region Proposal Network (RPN) \cite{ren2015faster} to detect multiple organs in CT volumes. However, their solution is not directly applicable to 3D fetal US volumes, which has more challenges in image quality and subject orientations. \par
	In this work, we build a 3D RPN-based object detection framework for landmark detection in 3D US. Specifically, we firstly implement the one-stage 3D RPN as the backbone to achieve a balance between accuracy and efficiency. Then we exploit the recently introduced IoU-balanced classification loss \cite{wu2019iou} to improve the landmark detection without adding any computation cost. Finally, we propose a simple yet effective distance based graph prior to regularize the learning process. Experimental results show that the proposed method can localize the landmarks accurately and efficiently. \par
	
	
	\section{Methodology}
\begin{figure}[htb]
		\centering
		\includegraphics[width=1.0\linewidth]{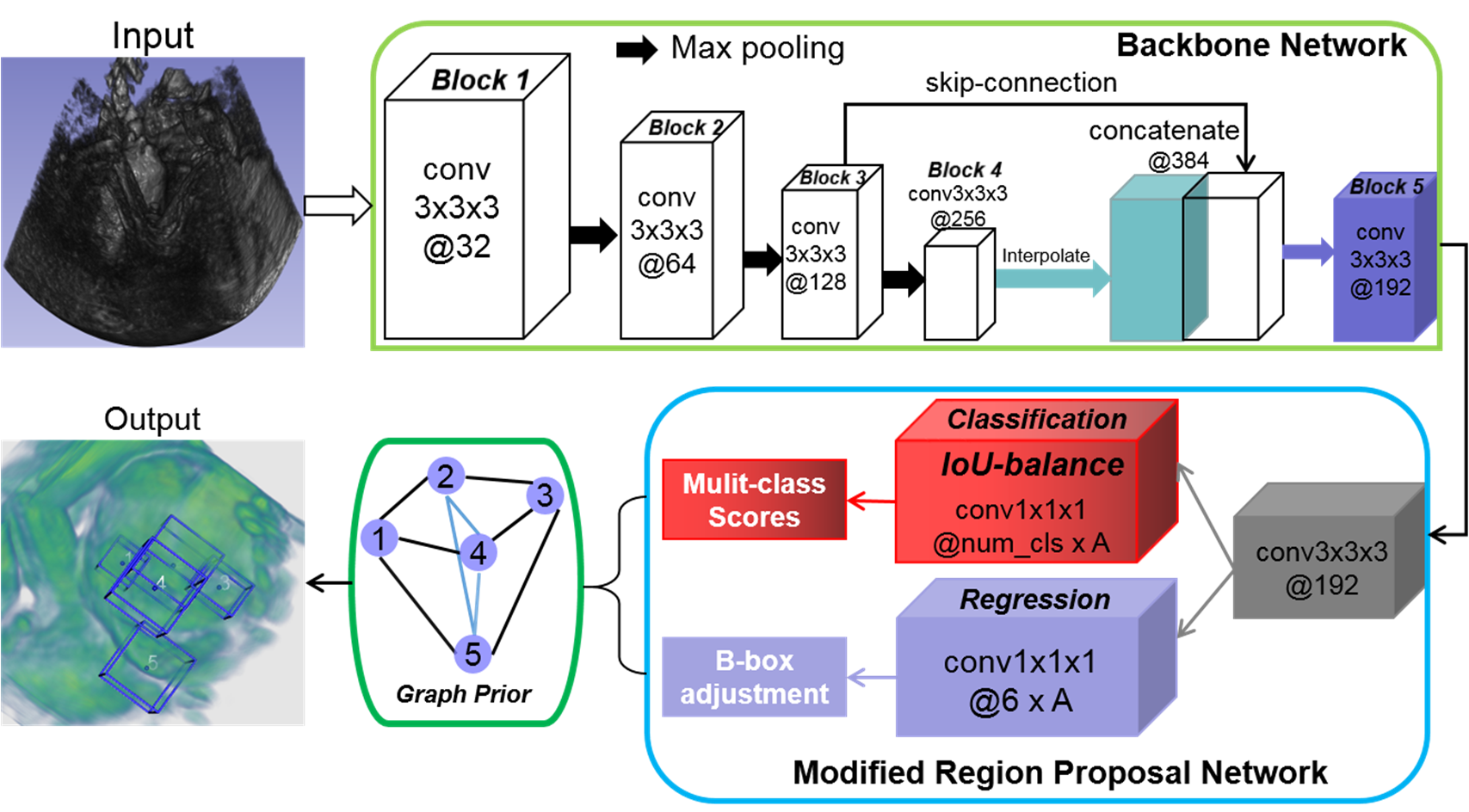}
		\caption{Framework of our method. ``conv$3 \times 3 \times 3$@32" means a layer with $3\times3\times3$ convolutional kernel and output channel 32. Each block has two convolution layers. ``A" is the anchor number in each cell.}
		\label{fig:framework}
		\vspace{-0.3cm}	
	\end{figure}

	Fig. \ref{fig:framework} provides an overview of our method. An input US volume is firstly passed to a convolutional network (CNN) for feature extraction and then assigned with a set of B-box candidates with different sizes and shapes. The last feature map block is then fed into a modified 3D RPN to predict the multi-class scores and B-box adjustment parameters. IoU-balanced classification loss is introduced in the RPN classification branch. Finally, a distance based graph prior filters all the candidates and produces the final B-box for each landmark. Note that the whole framework is implemented in 3D manner to avoid losing any spatial information. \par
	\par
	\subsection{Backbone Network for feature extraction}
	To extract rich features for region proposal, our backbone uses a 3D CNN modified from VGG-16 \cite{simonyan2014very} (see Fig. \ref{fig:framework}). Note that after three down-sampling layers, the learned feature maps become coarser and are not suitable for pinpointing landmark locations. Inspired by the design of U-Net \cite{ronneberger2015u}, the model aggregates feature maps from different layers to hallucinate high-resolution feature maps. Specifically, it employs an interpolation to upsample the feature block 4 directly, and then concatenate the output with the feature block 3.  It is followed by a \begin{math}3\times 3\times 3\end{math} convolution layer to generate the final feature maps. Batch normalization and ReLu are applied universally. \par
	\subsection{Modified RPN with IoU-balanced Classification}
	The feature map generated by the backbone network is passed into a modified 3D RPN \cite{ren2015faster} to generate candidate B-boxes. As illustrated in Fig. \ref{fig:framework}, the 3D RPN consists of a \begin{math}3\times 3\times 3\end{math} convolution layers followed by two sibling \begin{math}1\times 1\times 1\end{math} convolution layers. Specifically, each 3D anchor is associated with \begin{math}K + 1\end{math} class scores, indicating the presence of \begin{math}K\end{math} target points and background objects in it. The 6 adjustment parameters \begin{math}(t_x, t_y, t_z, t_w, t_h, t_d)\end{math} modify the anchor in position and size to better fit the potential target B-box. As the target fetal facial landmarks tend to be small, we set the target B-box size for fetal left and right eyes to $14^{3}$ voxels and the size of middle eyebrow, nose tip and chin are $24^{3}$ voxels. We use 4 base sizes (i.e. $\left\lbrace 13, 16, 20, 28\right\rbrace$) in each spatial dimension to define 64 anchors at each feature map cell. The produced B-box candidates can be expressed as:\par
	\begin{footnotesize}
	\begin{equation}
		\begin{aligned}
			\left\{
			\begin{aligned}
				x & = x_a + t_xw_a       & \\
				y & = y_a + t_yh_a       & \\
				z & = z_a + t_zd_a		 &
			\end{aligned}
			\right.  
			\left\{
			\begin{aligned}
				w & = w_ae^{t_w}  \\
				h & = h_ae^{t_h}  \\
				d & = d_ae^{t_d}
			\end{aligned}
			\right.
		\end{aligned}
	\end{equation}
	\end{footnotesize}
	Where the tuple \begin{math}(x,y,z,w,h,d)\end{math} denotes a candidate B-box centered at \begin{math}(x,y,z)\end{math} with a size of \begin{math}(w,h,d)\end{math}. 
	The landmark coordinates can be derived using the center of the final box.\par
	The classification losses adopted by vanilla RPN is independent of the localization task, while the final B-box is only determined by classification score. This often results in a scenario where a B-box with high classification score may have large localization error. To address this, we modify the loss function by adding an IoU-balanced classification loss \cite{wu2019iou} in the RPN classification branch to enhance the task correlation. It can be formulated as:\par
	\begin{footnotesize}
	\begin{equation}
		L_{cls} = \sum_{i \in {Pos}}^{N}\textit{w}_i(IoU_i)\ast\textrm{CE}(p_i,\hat{p}_i) + \sum_{i\in Neg}^{M}\textrm{CE}(p_i,\hat{p}_i)
		\label{Equ.(2)}
	\end{equation}

	\begin{equation}
		\textit{w}_i(IoU_i) = {IoU}^\eta_i\dfrac{\sum_{i=1}^{n}\textrm{CE}(p_i,\hat{p}_i)}{\sum_{i=1}^{n}{IoU}^\eta_i\textrm{CE}(p_i,\hat{p}_i)}
		\label{Equ.(3)}
	\end{equation}
	\end{footnotesize}
	In our modified loss $L_{cls}$, the weights assigned to positive examples are correlated with the IoU between the regressed B-boxes and corresponding ground truth. As a result, the samples with high IoU are up-weighted and the ones with low IoU are down-weighted adaptively based on their IoU after a bounding box regression. Thus, the correlation between the classification scores and the localization accuracy are enhanced. In other words, the classification score of candidate boxs will match their localization regression better. In Eq.(3), the parameter \begin{math}\eta\end{math} can regulate IoU-balanced classification loss focus on the samples with high IoU and suppresses the ones with low IoU. In this work, \begin{math}\eta\end{math} is set as 1.75.

\subsection{Graph Prior Regularized B-box Filtering}
\begin{figure}[htb]
	\centering
	\includegraphics[width=1.0\linewidth]{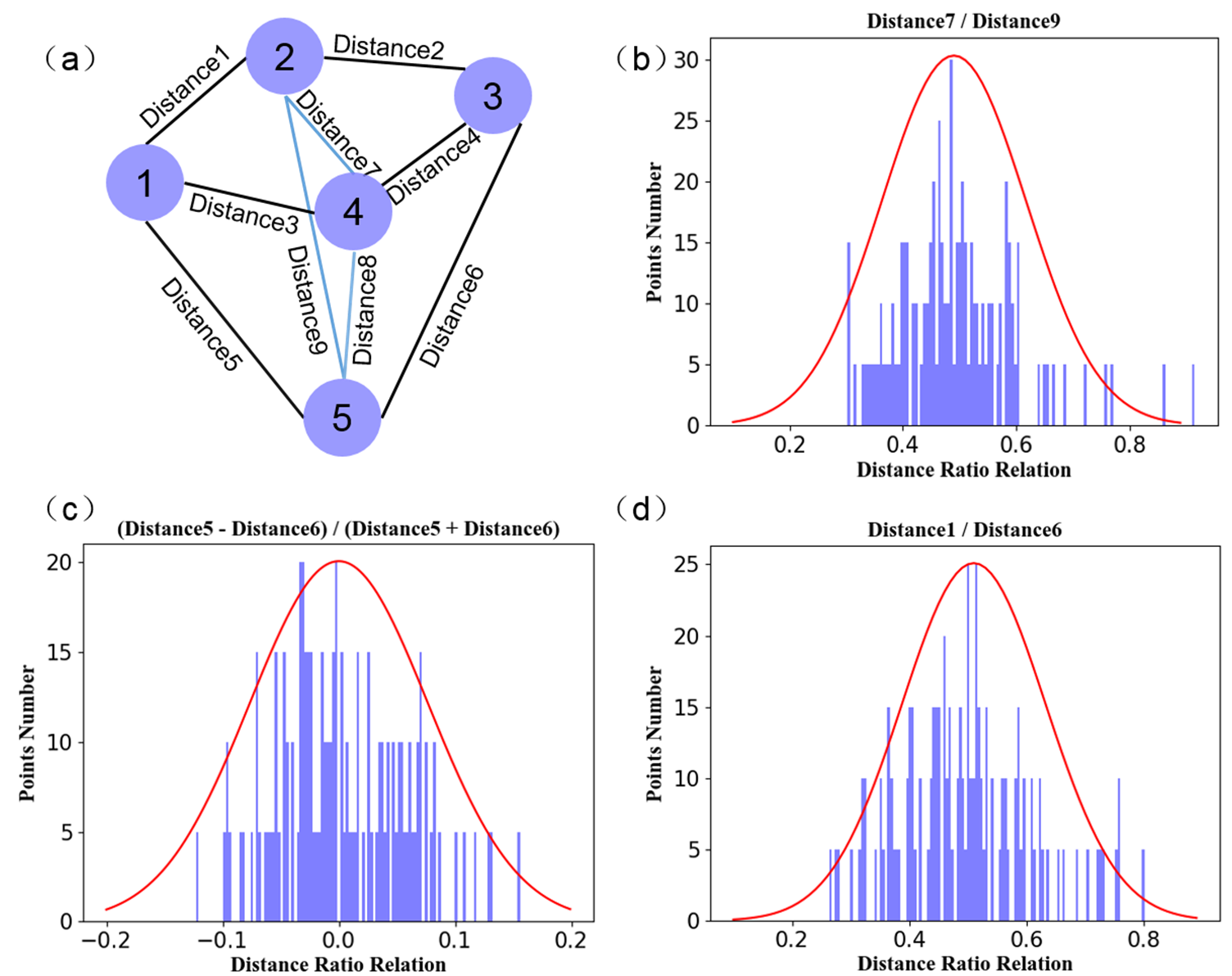}
	\caption{Graph prior of distance ratio. (a) Graph of 5 fetal facial landmarks, fetal left eye (node 1), middle eyebrow (node 2), right eye (node 3), nose (node 4) and chin (node 5). (b-d) Distribution of different distance ratios. Red curves are the fitted curves.}
	\label{fig:graph_prior}
	\vspace{-0.3cm}	
\end{figure}

In our modified RPN, a series of candidate B-boxes are generated for each landmark class, in which only the first two B-boxes with the highest classification scores for 5 landmarks are selected. These 10 B-boxes form 25 combinations of fetal facial landmarks. Outliers or false positives are contained in these combinations. We hence introduce a graph prior to filter out the best one from them. The graph prior makes full use of the anatomical prior. Different from previous heavy solutions \cite{tuysuzoglu2018deep,wang2019joint} in modelling the anatomical knowledge, our distance based graph prior is simple yet effective for landmark localization refinement. As shown in Fig. \ref{fig:graph_prior} (a), five landmarks of fetal face are modelled as the nodes of a graph with 9 edges. Graph edge is the Euclidian distance (\textit{Dist}) between two landmarks. Based on the graph, instead of directly using the Dist distribution as prior, we propose to adopt the distribution of Dist ratio between two edges for better scale-invariance and robustness. Specifically, for asymmetric edge $a, b$, we use the statistic distribution of $Dist(a)/Dist(b)$. For symmetric edge $c,d$, we use the statistic distribution of $(Dist(c)-Dist(d))/(Dist(c)+Dist(d))$. As shown in Fig. \ref{fig:graph_prior}(b-d), based on the training dataset, these ratios rougly follow different Gaussian distributions. \par

\par
Specifically, we use a strategy called anti-normal distribution penalty. When a ratio shifts from proper location, its probability value on the distribution curve will decrease. When the ratio is close to a reasonable range, it is likely to be at the peak of the curve. We directly calculate the reciprocal of the normal distribution of the distance ratio and then sum the multiple relationships:\par
\begin{footnotesize}
\begin{equation}
\begin{aligned}
L_{penalty} = \sum_{i=1}^{n}{\sigma_i}*\exp({\dfrac{(r_i-\mu_i)^{2}}{2{\sigma_i}^{2}}}),
\end{aligned}
\end{equation}
\end{footnotesize}
where \begin{math} r_i, \mu_i, \sigma_i \end{math} are the ratio value, distribution mean and variance, repectively. For the ratio that deviate from the mean, its penalty value is large, while when the point is close to the mean, the penalty value is small. The combination with the lowest $L_{penalty}$ value is treated as our final prediction result.

\section{Experimental Results}
\label{sec:typestyle}
In this study, we comprehensively evaluate our method on the 3D US fetal facial data, consisting of 152 US volumes with different postures. Voxel resolution is $1.3\times1.3\times1.3 mm^3$. 32 of 152 volumes are randomly selected as the testing set and the rest as the training set. Data augmentation, including rotation, flipping, is used. Random rotation is between $-25^{\circ}$ and $25^{\circ}$. In the experiment of landmark detection based on object detection, we specify an anchor as positive if it has the highest IoU with the ground truth or its IoU with ground truth is above 0.5. An anchor is considered as negative if its IoU with every ground truth is less than 0.25. For the anchors having 0.25 $\leq$ IoU $<$ 0.5 with any ground truth, they are not considered in this work. Because we use the target detection method to do landmark detection, reported evaluation metrics consist of the average distance between predicted and ground-truth landmarks($\textrm{d-mean$[mm]$}$) along with standard object detection metrics, including {$\textrm{AP}$} (averaged on IoUs from 0.5 to 0.95), {$\textrm{AP}_{35}$} (AP for threshold 0.35), {$\textrm{AP}_{75}$} (AP for threshold 0.75) and {$\textrm{mIoU}$} (the mean IoU across all categories). \par
After IoU-balanced classification correction, those candidate boxes with higher IoU will get higher classification scores. We then use prior knowledge to optimize the distance distribution between points to refine the detection results. Table \ref{table1} shows quantitative comparison between the proposed framework (IG-RPN, `IG' denotes IoU-balance loss and graph prior) and other typical object detection methods.
\begin{table}[!htbp]
	\centering
	\caption{Quantitative comparison among methods.} 
	\scriptsize
	\renewcommand\tabcolsep{4.5pt}
	\begin{tabular}{c|cccccccc}
		\hline 
		Method&$\textrm{AP}$ &$\textrm{AP}_{35}$&$\textrm{AP}_{50}$&$\textrm{AP}_{75}$&$\textrm{mIoU}$&$\textrm{d-mean}$&$\textrm{time(ms)}$ \\ 
		\hline
		Faster R-CNN&80.63&94.38&81.25&26.87&62.87&4.74&575 \\
		IG-SSD &80.00&92.50&80.00&22.50&61.46&5.21&449\\ 
		Heatmap&-&-&-&-&-&5.30&1100 \\
		IG-RPN &\textbf{82.50}&\textbf{95.00}&\textbf{82.50}&\textbf{27.50}&\textbf{63.77}&\textbf{4.38}&515\\ 
		\hline
	\end{tabular}
	\label{table1}
\end{table}
\begin{table}[htbp]
	\centering
	\caption{Ablation study of IoU-balance loss and graph prior on AP in different IoU threshold.}
	\scriptsize
	\renewcommand\tabcolsep{3.3pt}
	\begin{tabular}{cc|c|ccccccc}
		\hline
		IoU   &  Prior  & \multirow{2}*{Method} & \multirow{2}*{$\textrm{AP}$} & \multirow{2}*{$\textrm{AP}_{35}$} & \multirow{2}*{$\textrm{AP}_{50}$} & \multirow{2}*{$\textrm{AP}_{75}$} & \multirow{2}*{$\textrm{mIoU}$} & \multirow{2}*{$\textrm{d-mean}$}&$\textrm{time}$ \\
		balance &  Know   &                       &                              &                                   &                                   &&&&$\textrm{(ms)}$\\ \hline
		&         &  \multirow{4}*{SSD\cite{liu2016ssd}}   &            78.12             &               90.00               &               78.75               &               20.62               &             60.29             &               5.42&419               \\
		$\surd$ &         &                       &            78.12             &               91.25               &               78.12               &               22.50               &             61.31              &               5.38  &419             \\
		& $\surd$ &                       &            77.50             &               90.62               &               77.50               &          \textbf{22.88 }          &             61.10             &               5.28&449               \\
		$\surd$ & $\surd$ &                       &        \textbf{80.00}        &          \textbf{92.50}           &          \textbf{80.00}           &               22.50               &        \textbf{61.46}         &          \textbf{5.21} &449          \\ \hline
		&         &  \multirow{4}*{RPN}   &             78.12             &               95.00               &               78.12               &               24.37                &             62.24              &               4.77 &484              \\
		$\surd$ &         &                       &             81.25             &               95.00                &               81.87                &               26.87                &             63.54              &               4.43 &484              \\
		& $\surd$ &                       &             81.25             &           \textbf{95.63}           &               81.88                &           \textbf{28.13}           &             63.74              &               4.52 &515               \\
		$\surd$ & $\surd$ &                       &        \textbf{82.50}         &               95.00                &           \textbf{82.50}           &               27.50                &         \textbf{63.77}         &           \textbf{4.38} &515          \\ \hline		
	\end{tabular}
	\label{table2}
\end{table}\par
In Table \ref{table1}, we compare our model with other typical methods in landmark detection. Faster R-CNN and SSD are classic object detection framework. For this experiment, we replace the detection part of our model with them and get evaluation metrics in the first two rows of Table \ref{table1}. Our model outperforms other detection-based networks on both average distance and object detection metrics. Moreover, our method is superior to the average distance and efficiency when compared with the heat map point regression method.\par

\begin{figure}
	\centering
	\includegraphics[width=1.0\linewidth]{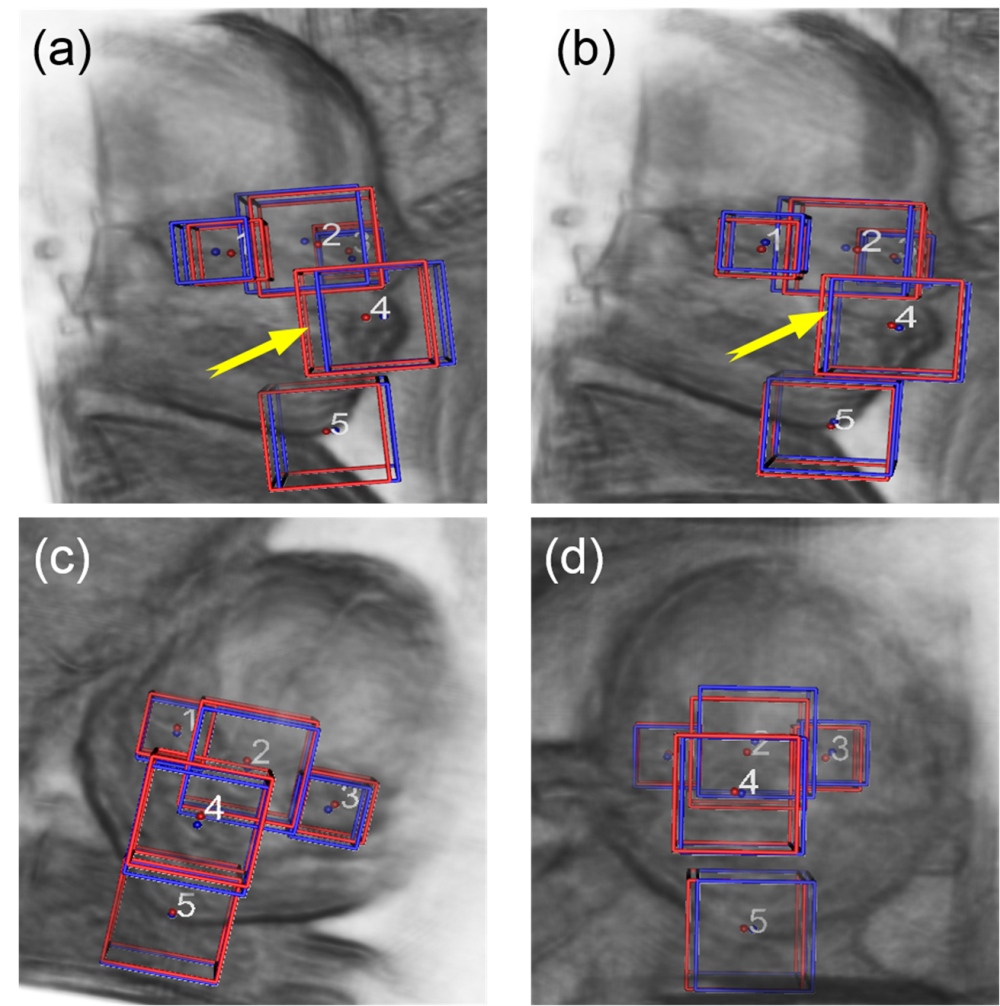}
	\caption{Visualization results. (a) and (b) are the result from RPN and our IG-RPN, respectively. (c) and (d) are another two results of our IG-RPN. Red box is the landmark ground truth, blue box is the prediction result.}
	\label{fig:visualization}
\end{figure}

Table \ref{table2} shows the ablation study on our proposed IoU-balance loss and graph prior on the original results under different detection methods. For the SSD, the IoU-balanced classification loss and graph prior knowledge can improve mIoU by 1.0$\%$ and 0.8$\%$, respectively. Combining them can improve {$\textrm{mIoU}$} by 1.2$\%$, and can improve {$\textrm{AP}_{35}$}, {$\textrm{AP}_{50}$} and {$\textrm{AP}_{75}$} by 2.5$\%$, 1.3$\%$ and 2.0$\%$, respectively. For our RPN, the IoU-balanced classification loss and the graph prior knowledge can both improve AP by 3.1$\%$, and combing them can improve AP by 4.3$\%$. For {$\textrm{AP}_{50}$}, {$\textrm{AP}_{75}$} and {$\textrm{mIoU}$}, the ensemble refinement method can improve them by 4.4$\%$, 3.1$\%$ and 1.5$\%$, respectively. We believe that these two refinement schemes are complementary, so we can achieve better results when they are united, especially the significant improvements on {$\textrm{AP}_{50}$} and {$\textrm{AP}_{75}$}. Our detection results and predicted landmarks are visualized in Fig. \ref{fig:visualization} to illustrate the importance of IoU balance loss and graph prior knowledge. Comparing Fig. \ref{fig:visualization} (a) and Fig. \ref{fig:visualization} (b), our method makes the IoU of the 4th point larger and the centroid distance smaller.\par
\section{conclusions}
In this paper, we propose to localize fetal facial landmarks in US volumes from the object detection perspective. As a highlight of our work, IoU-balance loss enhances the correlation between classification and localization. Our graph prior efficiently exploits anatomical knowledge and further filters the landmark candidates for refinement. Results show that, the proposed method outperforms some of the state-of-the-art methods in both efficacy and efficiency.\par



\bibliographystyle{IEEEbib}
\bibliography{refs}

\end{document}